\documentclass{article}

\usepackage[nonatbib, preprint]{neurips_2019}

\usepackage[backend=biber]{biblatex}
\usepackage[utf8]{inputenc} 
\usepackage[T1]{fontenc}    
\usepackage{hyperref}       
\usepackage{url}            
\usepackage{booktabs}       
\usepackage{amsfonts}       
\usepackage{nicefrac}       
\usepackage{microtype}      

\usepackage{graphicx}

\graphicspath{{./Figures/}} 
\usepackage{wrapfig} 

\usepackage[table,xcdraw]{xcolor} 

\usepackage{algorithm}
\usepackage{algorithmic}

\usepackage[textsize=tiny,disable]{todonotes}
\usepackage[textsize=tiny]{todonotes}

\usepackage{amsfonts}
\usepackage{mathtools}

\newtheorem{theorem}{Theorem}
\newcommand{\cN}{\mathcal{N}}
\newcommand{\cD}{\mathcal{D}}
\newcommand{\cH}{\mathcal{H}}

\newcommand{\cF}{\mathcal{F}}

\newcommand{\R}{\mathbb{R}}

\newcommand{\Ev}{\mathbb{E}}

\newcommand{\cL}{\mathcal{L}}

\newcommand{\cI}{\mathcal{I}}

\newcommand{\bs}{\boldsymbol}

\newcommand{\bF}{\bs F}

\newcommand{\DKL}{\cD_{\textnormal{KL}}}
\newcommand{\KL}{\mathcal{D}_\textnormal{KL}}

\newcommand{\sss}{\scriptscriptstyle}

\usepackage{verbatim}

\title{Stein Variational Online Changepoint Detection with Applications to Hawkes Processes and Neural Networks}

\author{%
  Gianluca Detommaso \thanks{Corresponding author: Gianluca Detommaso} \\
  InfoSec division \\
  G-Research \\
  London (UK)\\
  \texttt{detommaso.gianluca@gmail.com}\\
  \And
  Hanne Hoitzing\\
  InfoSec division \\
  G-Research \\
  London (UK) \\
  \texttt{hanne.hoitzing@gresearch.co.uk} \\
   \And
   Tiangang Cui\\
   School of Mathematical Sciences \\
   Monash University \\
   Melbourne (Australia) \\
   \texttt{tiangang.cui@monash.edu}\\
   \And
   Ardavan Alamir \\
   InfoSec division \\
   G-Research \\
   London (UK) \\
   \texttt{ardavan.alamir@gresearch.co.uk} \\
}

\begin{document}

\maketitle

\begin{abstract}
Bayesian online changepoint detection (BOCPD) \cite{adams2007bayesian} offers a rigorous and viable way to identify changepoints in complex systems. In this work, we introduce a Stein variational online changepoint detection (SVOCD) method to provide a computationally tractable generalization of BOCPD beyond the exponential family of probability distributions. We integrate the recently developed Stein variational Newton (SVN) method \cite{detommaso2018stein} and BOCPD to offer a full online Bayesian treatment for a large number of situations with significant importance in practice. We apply the resulting method to two challenging and novel applications: Hawkes processes and long short-term memory (LSTM) neural networks. In both cases, we successfully demonstrate the efficacy of our method on real data.
\end{abstract}

\section{Introduction}\label{sec:intro}
In most applied sciences and real-life scenarios, the ability to promptly detect and react to sudden changes is extremely desirable. Examples of current applications include hedge coverage in financial trading, attack detection in cybersecurity, prediction of natural disasters, and many others. In statistical analysis, the attempt to identify these changes is called \textit{changepoint detection}. 

Methods that fall under this category try to simultaneously minimize the following three important metrics: i) false negative rate, ii) false positive rate and iii) detection delay. 
False negatives must be avoided: missing the occurrence of an earthquakes could be fatal for thousands of people.
Similarly, avoiding false positives has significant importance: too many alerts will hide `true’ changes, leading the analyst to underestimate important information and to lose confidence in the statistical methodology. Finally, most applications require a realtime reaction once new data is observed. Online algorithms should minimize detection delay, without undermining the first two metrics. 

Among the literature in changepoint detection, probabilistic approaches have gained popularity for their ability to predict both the next observation and its uncertainty in an online fashion. 
A probabilistic approach that has significantly characterized the field 
is \textit{Bayesian online changepoint detection} (BOCPD) \cite{adams2007bayesian}. In its original formulation, this method exploits conjugate priors to construct predictive models in closed form. 
Although conjugate priors constitute an ingenious tool to decrease detection delay, their simplicity also represents their major limitation: if the data substantially differ from the simple model in use, false negative and false positive rates are very large.

Several Bayesian inference methods have been proposed to extend BOCPD to non-conjugate scenarios \cite{niekum2015online, mavrogonatou2016sequential, saatcci2010gaussian, turner2013online}. Although these methods form major contributions and have their own strengths, they also come with natural weaknesses. 
In this work, we propose a \textit{Stein variational online changepoint detection} (SVOCD) method, a combination of BOCPD and Stein variational inference \cite{liu2016stein}. Stein variational inference is a cutting-edge Bayesian inference methodology which transports, sequentially and deterministically, a set of particles towards a posterior probability density. The advantage of SVOCD compared to the extensions mentioned above is twofold: i) rather than merely approximating the posterior density, the empirical density represented by the particles asymptotically converges to the posterior density as the number of particles increases \cite{liu2017stein}, ii) rather than re-computing the posterior density from scratch when new data become available, it can be updated quickly which is crucial for online applications. Quick updates are possible because the posterior density can be used as the initial particle density to infer the next posterior density as new data points arrive. Assuming the true posterior does not change significantly, the particle locations can be adjusted with a few iterations. As a Stein variational algorithm we adopt Stein variational Newton (SVN), which was shown to drastically improve convergence speed and scalability to high-dimension compared to Stein variational gradient descent (SVGD) \cite{detommaso2018stein}.

Additionally, we successfully apply our methodology to two complex models which represent a large number of real-life scenarios and currently lack a rigorous Bayesian changepoint analysis:\footnote{Open source code is available at \href{https://github.com/gianlucadetommaso/Stein-variational-samplers}{gianlucadetommaso/Stein-variational-samplers.}.} i) Hawkes processes, and ii) the combination of BOCPD with a Bayesian \textit{long short-term memory} (LSTM) neural network model. Hawkes processes are an example of point processes for which the intensity increases with the occurrence of an event and exponentially decays over time; they have been used in a wide range of applications \cite{embrechts2011multivariate, bacry2015hawkes, ogata1998space, zhou2013learning}. 
To the best of our knowledge, this is the first attempt to perform online changepoint detection on Hawkes process in a fully Bayesian fashion. 
In many situations, no model exists to describe the evolution of data. LSTMs form a flexible modelling tool which can be trained to describe a sequence of data points and predict future points. However, the absence of an explicitly defined model structure can lead to large computational costs: LSTM's descriptive power comes from over-parametrization, making training computationally intensive and likely to end up in flat regions in parameter space. In this paper, we try to overcome these issues by combining a Bayesian formulation of LSTM with BOCPD and train the model using SVN.


The paper is structured as follows: Section \ref{sec:SVOCD} describes the background of BOCPD and SVN, and then presents SVOCD. 
In Sections \ref{sec:Hawkes} and \ref{sec:LSTM}, we apply SVOCD to Hawkes processes and LSTM, respectively. A conclusion is stated in section \ref{sec:conclusion}.

\section{Stein variational online changepoint detection}\label{sec:SVOCD}
In this section, we introduce Stein variational online changepoint detection (SVOCD). SVOCD generalizes BOCPD to probability distributions beyond the exponential family by using the Stein variational Newton method to perform online inference. 

\subsection{Background on Bayesian online changepoint detection}\label{sec:BOCPD}

Suppose we sequentially observe data points $y_{1:m}$, where the subscript denotes the observation time. Assuming that each observation $y_i$ depends on a model driven by a hidden parameter $\theta\in \R^d$, changepoint detection aims to identify abrupt changes in the parameter $\theta$. We denote a \textit{changepoint} by a time index $\tau > 1$ at which the abrupt change in $\theta$ occurs.
We will focus on online changepoint detection: given past observations $y_{1:m}$, we want to detect whether $\theta$ at time $m+1$ is the same as $\theta$ at time $m$. 
We want to perform this task recursively as new data becomes available.

Bayesian online changepoint detection (BOCPD) has been introduced as a probabilistic approach for online detection of changepoints in a time series \cite{adams2007bayesian}. The algorithm has pioneered a considerable amount of interesting follow-up work. Here we provide a description of the general formulation of BOCPD.\footnote{
We adopt a formulation without the concept of \textit{run length}, however the method is equivalent.}
BOCPD adopts the following reasonable assumption.

\textbf{A1.} Observed data before and after changepoints are independent. That is, $y_i$ is independent of $y_j$ if there exists a changepoint $\tau$ such that $i < \tau \leq j$. This way, the dynamics of the underlying system after a changepoint is not affected by what happened before the changepoint. 

Let us define $\tau_{m+1} \in \{1,\dots, m+1\}$ to be the changepoint indicator at time $m+1$ which records the time of the occurrence of the last changepoint. The case $\tau_{m+1}=1$ indicates there has been no changepoint up until time $m+1$. Although a priori $\tau_{m+1}$ can assume any value between 1 and $m+1$, in practice one should consider pruning the possible set of changepoints according to their posterior probability for significant computational speed-ups \cite{turner2009adaptive}.

\paragraph{Predictive posterior.} Suppose we have observed $y_{1:m}$ and we want to detect whether $y_{m+1}$ is a changepoint. For this purpose, we introduce the predictive posterior density $p(y_{\sss m+1}|y_{\sss 1:m})$, which measures the probability that $y_{m+1}$ is observed given $y_{1:m}$. However, because of assumption \textbf{A1}, $y_{m+1}$ is only dependent on observations since the last changepoint $\tau_{m+1}$. Then, if we define $Y_{\tau_{m+1}}\coloneqq \{\tau_{m+1},y_{\tau_{m+1}:m}\}$ to be the information set given by both the changepoint $\tau_{m+1}$ and the sequence of observed data points $y_{\tau_{m+1}:m}$ (we define $y_{m+1:m}=\emptyset$), we can marginalize the predictive posterior density as follows:
\begin{equation}\label{eq:predPost2}
p(y_{\sss m+1}|y_{\sss 1:m}) = \hspace{-2pt}\sum\nolimits_{\sss \tau_{m+1}=1}^{\sss m+1} p(y_{\sss m+1}|Y_{\sss \tau_{\sss m+1}})\,p(\tau_{\sss m+1}|y_{\sss 1:m})\,.
\end{equation} 
We will now analyse the two factors on the right-hand-side of equation \eqref{eq:predPost2}.

\paragraph{Predictive model.} $p(y_{\sss m+1}| Y_{\sss \tau_{\sss m+1}})$ denotes the predictive probability given the last changepoint $\tau_{m+1}$. By marginalising $y_{m+1}$ over the hidden parameter $\theta$, we can write
\begin{equation}\label{eq:predGivenTau}
\!\!\!\! p(y_{\sss m+1}|Y_{\sss \tau_{\sss m+1}}) = \int p(y_{\sss m+1}|Y_{\sss \tau_{m+1}}, \theta)\,p(\theta | Y_{\sss \tau_{\sss m+1}})\,d\theta,
\end{equation}
where $p(\theta | Y_{\sss \tau_{\sss m+1}})$ denotes the posterior distribution of $\theta$ and we refer to $p(y_{\sss m+1}| Y_{\sss \tau_{m+1}}, \theta)$ as the \textit{predictive likelihood}. 

In original BOCPD, the authors exploit conjugate priors for exponential families of probability distributions to express the predictive density \eqref{eq:predGivenTau} in closed form. 
In section \ref{sec:SVOCD}, we will generalize BOCPD to non-exponential families of probability distributions by introducing Stein variational Newton. This enables us to accurately approximate $p(y_{\sss m+1}|Y_{\sss \tau_{\sss m+1}})$ for more complex model choices that can better represent the data and their changepoints, while keeping the detection delay small. 

\paragraph{Changepoint posterior.} $p(\tau_{\sss m+1}|y_{\sss 1:m})$ denotes the posterior probability of the changepoint indicator $\tau_{m+1}$. Using an approach analogous to \cite{adams2007bayesian}, it is easy to show that the joint probability of $\tau_{m+1}$ and $y_{1:m}$ can be recursively expressed as
\begin{equation}\label{eq:recJoint}
\!\!\!\! p(\tau_{\sss m+1}, y_{\sss 1:m})\hspace{-3pt} =\hspace{-6pt} \sum\nolimits_{\sss \tau_m = 1}^{\sss m} \!\! p(y_{\sss m}|Y_{\sss \tau_{\sss m}})\,p(\tau_{\sss m+1}|\tau_{\sss m})\,p(\tau_{\sss m},y_{\sss 1:m-1}).\!\!\!
\end{equation}
Hence, the joint density on the left-hand-side of \eqref{eq:recJoint} can be evaluated by a forward message-passing algorithm which stores the joint density evaluations at the previous iteration and updates them accordingly. The posterior density can then be recovered by normalizing the joint density via $p(y_{\sss 1:m}) = \sum_{\tau_{\sss m+1}=1}^{\sss m+1} p(\tau_{\sss m+1},y_{\sss 1:m})$. Note that, given $\tau_m$, we can only have either $\tau_{m+1} = \tau_m$ if $y_{m+1}$ follows the same dynamics as $y_{m}$, or $\tau_{m+1} = m+1$ if $m+1$ is a changepoint. Then, we define the \textit{changepoint prior} density $p(\tau_{m+1}|\tau_m)$ as being equal to either $H_m$ (if $\tau_{m+1} = m+1$) or $1 - H_m$ (if $\tau_{m+1} = \tau_m$), where $H_m$ can be interpreted as a \textit{hazard rate}. Hence, whenever $\tau_{m+1} \ne m+1$, the sum over $\tau_m$ in \eqref{eq:recJoint} reduces to a single term with $\tau_m = \tau_{m+1}$.

\subsection{Background on Stein variational Newton}\label{sec:SVN}
Consider an intractable target density $\pi$ on $\R^d$ that we wish to approximate via an empirical measure or, equivalently, a collection of particles. Given a set of particles $(\theta^{(k)})_{k=1}^{N_\theta}$ characterizing an initial reference density $q_0$, we seek a transport map $T:\R^d\to \R^d$ such that $T_*q_0$, the push-forward map of $q_0$ through $T$, is a close approximation of $\pi$.\footnote{If $T$ is an invertible map, the push-forward map is defined by $T_*q(\theta) = q(T^{-1}(\theta))\,|\det(\nabla_\theta T(\theta))|$.} Such a map $T$ is not unique: there exist an infinite number of such maps that can serve the purpose \cite{villani2008optimal}. In the following, we construct $T$ as a composition of simple maps $T_l$ which are iteratively applied on reference densities $q_l$ such that $q_{l+1} = T_{l*}q_l$. We define each $T_l$ as a perturbation $Q_l$ of the identity map:
\begin{equation}\label{eq:pertIdentity}
T_l(\theta) = \theta + Q_l(\theta)\,.
\end{equation}
When applied to the current reference density $q_l$, equation \eqref{eq:pertIdentity} defines the push-forward measure $T_{l *} q_l$ as an update of $q_l$ itself along the direction $Q_l$. The latter will be taken along a vector-valued \textit{Reproducing Kernel Hilbert Space} (RKHS) $\cH^d\simeq \cH\times \cdots \times \cH$ characterized by a kernel $k(\cdot,\cdot)$.

\paragraph{A variational approach.} We define the functional
\begin{equation}\label{eq:objective_KL}
Q \mapsto J_{q_l}[Q] \coloneqq  \KL((I+Q)_*\,q_l\,||\,\pi)\,,
\end{equation}
with $Q\in \cH^d$. $J_{q_l}[Q]$ measures the Kullback-Leibler (KL) divergence $\DKL$ between the push-forward map of $q_l$, along the direction $Q$, and $\pi$. Thus, we want to find a map $Q_l$ such that $J_{q_l}[Q_l]<J_{q_l}[\bs 0]$, where $\bs 0(\theta) = 0$ denotes the zero map. In other words, we are constructing a sequence of densities $q_0,q_1,q_2, \dots$ that weakly converges to $\pi$ (see \cite{liu2017stein} for convergence results).

It was shown in \cite{liu2016stein, detommaso2018stein} how to define a functional gradient $\nabla J_{q_l}[\bs 0]$ and functional Hessian $\nabla J_{q_l}[\bs 0]$ of the map in \eqref{eq:objective_KL}, where $\bs 0$ is the null map, which symbolize the evaluation of the variational information at the current density $q_l$. For details about the methodology we refer to \cite{liu2016stein, detommaso2018stein}. Here we report the following theorem.

\begin{theorem}\label{thm:SVN}
	With the notation above, we have
	\begin{align}
	\hspace{-0.3cm} \nabla J_{q_l}[\bs 0](\phi)&  = -\Ev_{\theta \sim q_l}[\nabla_\theta\log\pi(\theta)k(\theta,\phi)
	+ \nabla_\theta k(\theta, \phi)]\,, \label{eq:nabla1}\\
	\hspace{-0.3cm} \nabla^2 J_{q_l}[\bs 0](\phi, \psi) & = \Ev_{\theta \sim q_l}[-\nabla_\theta^2\log\pi(\theta)k(\theta,\phi)k(\theta,\psi)
	 + \nabla_\theta k(\theta, \phi)\nabla_\theta k(\theta,\psi)^\top]\,.\label{eq:nabla2}
	\end{align}
\end{theorem} 
Given the variational informations in \eqref{eq:nabla1} and \eqref{eq:nabla2}, $Q_l$ is constructed via a Newton-type iteration (see the supplementary material for details). The overall method is addressed as \textit{Stein variational Newton} (SVN) and a possible implementation is described in Algorithm 1. 

\subsection{A new method: BOCPD via SVN}\label{sec:BOCPDplusSVN}
\begin{wrapfigure}{t}{0.53\textwidth}
		 \includegraphics[trim = 0cm 9cm 0cm 2cm, width=0.53\textwidth]{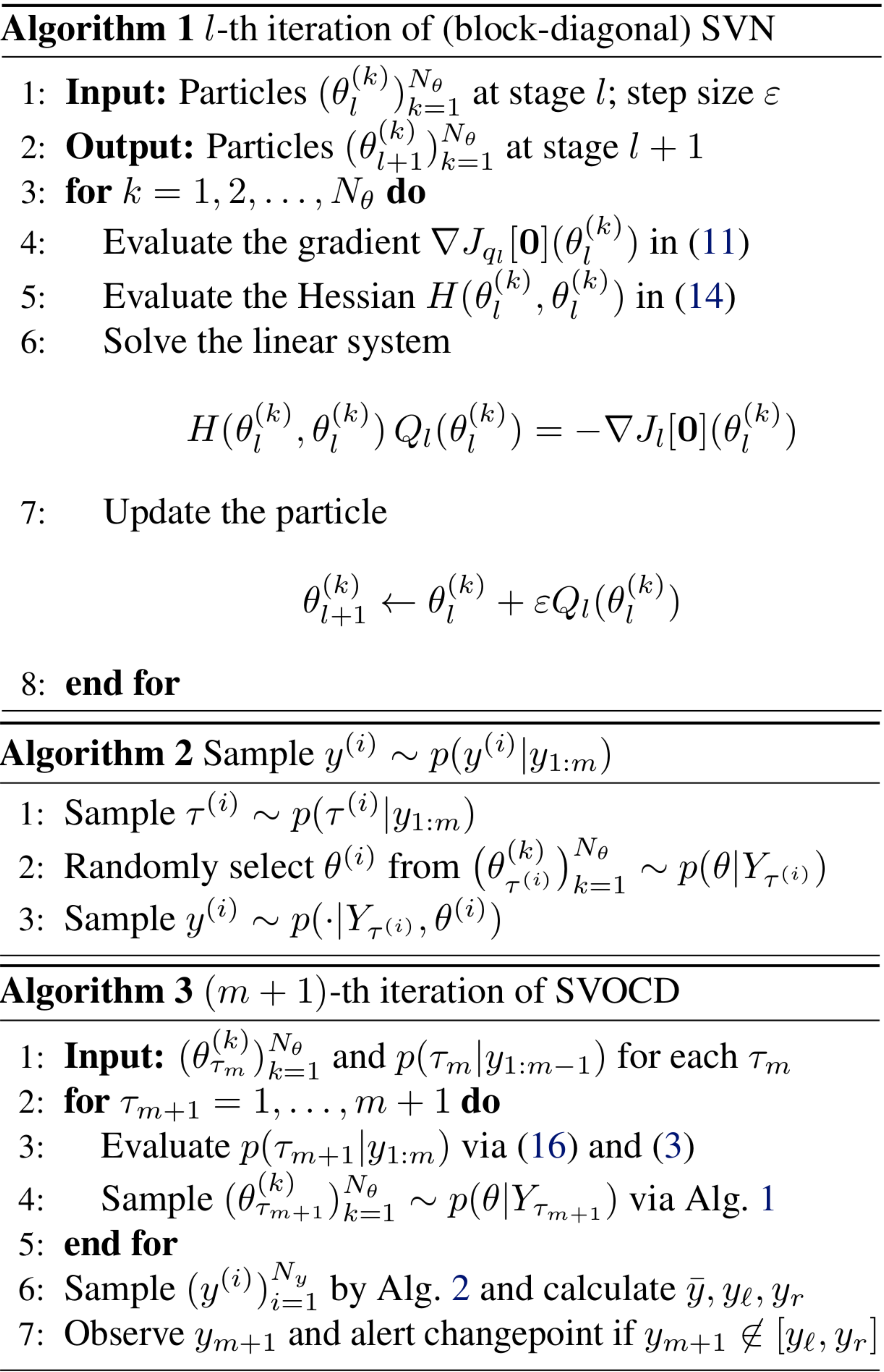}
		\label{alg:SVOCD}
\end{wrapfigure}

Here we introduce a novel method: Stein variational online changepoint detection (SVOCD). This algorithm generalizes BOCPD to non-exponential families of probability distributions. The $(m+1)$-th iteration of SVOCD is described in Algorithm \ref{alg:SVOCD}, which we break down into the following steps.

\paragraph{Changepoint posterior update (line 3).} Given samples $(\theta^{(k)}_{\tau_{m}})_{k=1}^{N_\theta}\sim p(\theta|Y_{\tau_m})$ and the changepoint posterior $p(\tau_{m}|y_{1:m-1})$ from the previous iteration, this step aims to update the changepoint posterior by the recurrent relation in \eqref{eq:recJoint} given the new observation $y_m$. We observe that this involves the evaluation of the posterior probability $p(y_m|Y_{\tau_m})$, which is a not available explicitly for a non-exponential family of probability distributions. However, because the samples $(\theta^{(k)}_{\tau_{m}})_{k=1}^{N_\theta}$ are available to us, we can simply estimate it by the Monte Carlo approach
\begin{equation}\label{eq:mc_evidence}
p(y_{m}|Y_{\tau_{m}}) \approx \frac{1}{N_\theta}\sum\nolimits_{k = 1}^{N_\theta} p(y_{m}|Y_{\tau_{m}}, \theta_{\tau_{m}}^{(k)})\,.
\end{equation}\vspace{-1.5em}
\paragraph{Samples update (line 4).} Next, we use SVN to generate samples $(\theta^{(k)}_{\tau_{m+1}})_{k=1}^{N_\theta}\sim p(\theta|Y_{\tau_{m+1}})$. Note that, given the changepoint $\tau_{m+1}$, we can only have either $\tau_{m+1} = m+1$ in the case of a changepoint or $\tau_{m+1} = \tau_{m}$ otherwise.
In the case $\tau_{m+1} = m+1$, the information set $Y_{\tau_{m+1}}$ contains no data points and, as a consequence, the posterior distribution $p(\theta|Y_{\tau_{m+1}})$ corresponds to the prior $p(\theta)$. A collection of samples $(\theta_{\tau_{m+1}}^{(k)})_{k=1}^{N_\theta}$ can now simply be taken from this prior.
In the case $\tau_{m+1} = \tau_{m}$, we have $Y_{\tau_{m+1}} = Y_{\tau_{m}} \cup \{y_{m}\}$ which means the following decomposition holds:
\begin{equation}\label{eq:seqDecomp}
p(\theta|Y_{\tau_{m+1}}) \propto p(y_{\sss m}|Y_{\tau_{\sss m}}, \theta)\,p(\theta|Y_{\tau_{\sss m}})\,.
\end{equation}
The relation in \eqref{eq:seqDecomp} shows that we can recast $p(\theta|Y_{\tau_{m+1}})$ as a sequential update of the previous posterior $p(\theta|Y_{\tau_{m}})$ through the information given by $y_m$. The transport method of SVN very well fits this framework: SVN can be initialized using the current particles $(\theta_{\tau_m}^{(k)})_{k=1}^{N_\theta} \sim p(\theta|Y_{\tau_{m}})$, which are then adjusted to get  $(\theta_{\tau_{m+1}}^{(k)})_{k=1}^{N_\theta} \sim p(\theta|Y_{\tau_{m+1}})$. Since the initialization of the particles is optimal up to the available information $Y_{\tau_{m}}$, the algorithm most likely needs only a few iterations to converge, particularly if the amount of information that $y_{m+1}$ adds to $Y_{\tau_{m}}$ is small.

\paragraph{Data prediction (line 6).} Given the updated changepoint posterior and particles, Algorithm 2 is a standard valid mechanism \cite{lynch2007introduction} to produce samples $(y^{(i)})_{i=1}^{N_y}$ from the predictive posterior density $p(y|y_{1:m})$. We can use these samples to work out a prediction for the next observation $y_{m+1}$ and statistics summarizing the distribution, for example left and right quantiles $y_\ell$ and $y_r$ in the case of one-dimensional data.
\paragraph{Data classification (line 7).} Finally, the data $y_{m+1}$ is observed and immediately alerted as a changepoint if it does not belong to the credible interval $[y_\ell,y_r]$.

\paragraph{Remarks.} We stress that the loops over $\tau_{m+1}$ can be executed in parallel. In particular, parallelizing the samples update step is fundamental to massively speed up the algorithm. Additional steps for pruning the set of possible values of $\tau_{m+1}$ or for optimizing over the hyper-parameters could be added \cite{turner2009adaptive, wilson2010bayesian} to Algorithm \ref{alg:SVOCD}, but this goes beyond the scope of this paper.

\section{Application to Hawkes processes}\label{sec:Hawkes}

In this section, we apply SVOCD to Hawkes processes: common self-exciting point processes that play a central role in analysing time series in a range of applications such as telecommunications, epidemiology, and neuroscience. Though frequentist's methods have previously been developed \cite{li2017detecting, price2017statistical}, this is, to the best of our knowledge, the first fully-Bayesian online treatment of detecting changepoints in Hawkes processes.

\paragraph{Hawkes processes.} Unlike standard inhomogeneous Poisson processes, the intensity function of a self-exciting process directly depends on the occurrence of past events, which can "excite" the arrival of future events. In a Hawkes process, the rate of arrivals bursts whenever an event occurs, and decays over time. We denote the sequence $(y_k)_{k\ge 1}$ to be the arrival times of the process. Given 
$Y_{\tau_{m+1}}$, the rate of arrival of the next event $y_{m+1}$ can be described by the following \textit{conditional intensity}:
\begin{equation}\label{eq:hawkes_cond_intensity}
\lambda_{\tau_{m+1}}(t) \coloneqq \mu + \gamma\sum\nolimits_{\stackrel{y_k\in Y_{\tau_{m+1}}}{y_k < t}}e^{-\delta(t - y_k)}\,,
\end{equation}
where $t > 0$, $\mu > 0$ is the baseline intensity rate, $\gamma > 0$ represents how much the intensity bursts whenever an event occurs, and $\delta > 0$ represents the decay rate of the intensity function. When no event has arrived yet, a Hawkes process behaves like a homogeneous Poisson process with parameter $\mu$. We define $\theta\coloneqq [\mu,\gamma,\delta]^\top$ as the 3-dimensional vector collecting all parameters. More general definitions of Hawkes processes could be considered (e.g. marked Hawkes processes, power decay functions, ...) \cite{rizoiu2017tutorial}, but we restrict ourselves to the most common one.

Given that we observed the events in $Y_{\tau_{m+1}}$ within the time interval $(y_{\tau_{m+1}-1},y_m]$, it can be shown that the predictive likelihood function for the next event $y_{m+1}$ is given by
\begin{equation}
\!\!\!\! p(y_{m+1}|Y_{\tau_{m+1}}\!, \theta)\! =\! \lambda_{\tau_{m+1}}\!(y_{m+1})\, e^{\!-\Lambda_{\tau_{m+1}}\!\!\big( (y_m, y_{m+1}]\big)\!\!}\,, \!\!
\end{equation}
where $\Lambda_{Y_{\tau_{m+1}}}(\cI) \coloneqq \int_{\cI}\lambda_{Y_{\tau_{m+1}}}(t)\,dt$ is known as a \textit{compensator}, for some time interval $\cI$ \cite{rasmussen2018lecture}.
The likelihood function can then be explicitly defined as
\begin{equation}\label{eq:hawkes_llkd}
 p(y_{\tau_{m+1}:m}|\tau_{m+1}\!,\theta)\, = \hspace{-8pt} \,\,\,\,\,\,\prod\nolimits_{i=\tau_{m+1}}^m \!\!\!\!\! \lambda_{\tau_{m+1}}\!(y_i)\, e^{-\Lambda_{\tau_{m+1}}(\cI_{\tau_{m+1}})}\,,
\end{equation}
where $\cI_{\tau_{m+1}} \coloneqq (y_{\tau_{m+1}-1}, y_m]$.
In order to enforce positivity for each component of $\theta$, we impose a log-normal prior distribution, i.e.~ $\ln \theta \sim \cN(\mu_0, \sigma_0^2 I)$, where $\mu_0$ and $\sigma_0$ are hyper-parameters. Using Bayes' Theorem, we have
\begin{equation}\label{eq:postDistr}
p(\theta| Y_{\tau_{m+1}}) \propto p(y_{\tau_{m+1}:m}|\theta, \tau_{m+1})\, p(\theta)\,.
\end{equation}

\paragraph{A choice for the Hessian.}
In order to apply SVOCD to Hawkes processes, a positive definite approximation of the Hessian of the log-likelihood density, $\nabla^2_\theta \log p(y_{\tau_{m+1}:m}|\tau_{m+1},\theta)$, is required. We represent this approximation by the asymptotic Fisher information, shown \cite{reinhart2018review} to be given by 
\begingroup\makeatletter\def\f@size{9}\check@mathfonts
\def\maketag@@@#1{\hbox{\m@th\normalsize\normalfont#1}}
\begin{equation}\label{eq:hawkes_hess_L}
H_{\cL, \tau_{m+1}}\!(\theta) \,\coloneqq \sum\nolimits_{\sss y_i \in Y_{\sss\tau_{m+1}}}\hspace{-10pt}\nabla_\theta \log\lambda_{\sss\tau_{m+1}}\!(y_i)\nabla_\theta \log\lambda_{\sss\tau_{m+1}}\!(y_i)^\top.\!\!
\end{equation}
\endgroup
An approximation of the Hessian of the log-posterior density $\nabla_\theta^2 \log p(\theta|Y_{\tau_{m+1}})$ can then be given by
\begin{equation}\label{eq:Hpi}
H_{\pi, \tau_{m+1}}(\theta) = \sigma_0^2 I + H_{\cL, \tau_{m+1}}(\theta)\,.
\end{equation}
We note that when calculating the gradient $\nabla_\theta \log p(\theta|Y_{\tau_{m+1}})$, each $\nabla\log\lambda_{Y_{\tau_{m+1}}}(y_i)$ in \eqref{eq:hawkes_hess_L} needs to be evaluated and hence the calculation of $H_{\pi, \tau_{m+1}}(\theta)$ does not require additional operations. 


\paragraph{Validation via SMC.} To benchmark the performance of SVOCD, we will also employ BOCPD using Sequential Monte Carlo (SMC) with adaptive systematic resampling \cite{doucet2009tutorial} to update $\theta$ samples (line 4, Algorithm 3). The importance density is taken as the Laplace approximation of the posterior, i.e. a Gaussian centered at the MAP with covariance matrix the inverse of the Hessian evaluated at the MAP. We do not aim to face the difficult task of a rigorous performance comparison, but rather we introduce an alternative not requiring structural choices of proposal or approximating densities. 

\paragraph{Application: WannaCry cyber attack.}
WannaCry caught world headlines in May 2017 by infecting over 200,000 computers and causing damages worth at least in the hundreds of millions of dollars. 
In this section, we consider the packet capture traffic logs of the WannaCry spread through three computers in a test environment.\footnote{Data can be found here: \href{https://www.malware-traffic-analysis.net/2017/05/18/index2.html}{https://www.malware-traffic-analysis.net/2017/05/18/index2.html}.} The spread of the malware triggers a snowball effect of logs as each computer gets infected. In order to capture this self-exciting phenomenon, we employ a Hawkes process to model the log arrivals in time and perform online changepoint detection to efficiently detect when the three computers become infected.

The data contains 207 time observations. The prior distribution for $\ln \theta$ was deliberately chosen as an uninformative Gaussian with parameters $\mu_0=0$ and $\sigma_0^2=10$. As we look for sudden bursts of activity, we construct a one-sided credible interval by taking $y_r$ as the $95$th percentile and we signal a changepoint $m$ whenever $y_m > y_r$. The hazard rate in the changepoint prior and the number of predictive samples were fixed at $H_m = 100$ and $N_y = 100$, respectively.

\begin{figure}[h!]
	\begin{center}
		\centerline{ \includegraphics[trim = 0cm 0cm 1cm 0cm, width=1.0\textwidth, height = 5cm]{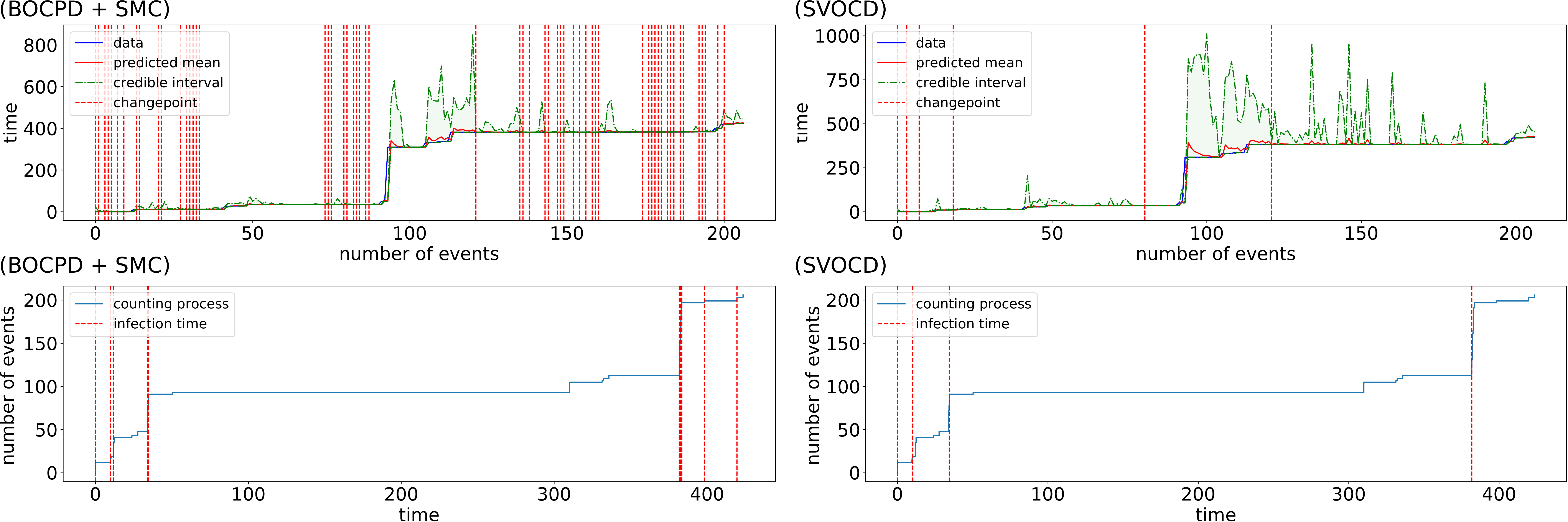}}
		\caption{\small{Hawkes model on WannaCry data. (left) BOCPD + SMC: several false positives are identified along the three infections. (right) SVOCD: after burn-in, only three changepoints are detected corresponding to the three infections. In the SVN algorithm, we used $N_\theta = 100$ particles and only 30 iterations. For BOCPD + SMC, 1000 particles were used.}}
		\label{fig:wannacry}
	\end{center}
	\vskip -0.2in
\end{figure}

Figure \ref{fig:wannacry} displays the results of both SVOCD as well as BOCPD in conjuction with SMC applied to the WannaCry data. In the top figures, 
the blue line represents the observations; vertical jumps indicate that no log events occur in that time interval, whereas horizontal regions indicate event arrivals close together. The red line is the average of the predictive distribution, attempting to reconstruct the data; the green shadowed area represents the credibility region up to the right $95$th percentile of the distribution; the vertical red lines are the detected changepoints. For the bottom figures we reverted the axes so that the blue line represents a counting process which increases by 1 every time an observation occurs.

\begin{wraptable}[]{r}{0.6\textwidth}
\begin{tabular}{ p{2.2cm} |p{1.6cm}| p{2.1cm}}
& SVOCD & BOCPD $\!$+ $\!$SMC \\
\hline

  False pos. rate & 1.03 (0.91) & 1.43 (0.99) \\
  False neg. rate & 0.37 (0.48) & 0.60 (0.49) \\
  \hline
  & SVN & SMC \\
  \hline
  MSE at $\tau = 15$ & $\sim 3 \times $10$^{-2}$ & $\sim 2 \times $10$^1$ \\
\end{tabular}
\caption{\small{Quantitative comparisons between SVOCD and BOCPD + SMC show lower false positive and false negative rates (mean (std)) for SVOCD (30 iterations). The MSE (as defined in the text) of SVN at an arbitrarily chosen timepoint is about 3 orders of magnitude lower than that of SMC when using 500 particles.}}\vspace{-0.5em}
\label{Tab: quantitative results}
\end{wraptable}

We find that SVOCD takes some time to adapt at the beginning of the time series, which is to be expected as the prior distribution is not yet properly tuned to the data. However, the algorithm quickly adapts and detects three meaningful changepoints. Thus, apart from the initial burn-in phase, all the detected changepoints correspond to drastic bursts in activity, i.e.~to the infections of the three computers, and no false positives were detected. For BOCPD + SMC, the algorithm keeps detecting changepoints without adapting to changes in data trends. Although these changepoints correspond to actual bursts in activity, several false positives are detected along with the machine infections.

\begin{wrapfigure}{r}{0.5\textwidth}
	\begin{center}
		\centerline{ \includegraphics[trim = 0cm 1cm 0cm 2cm, width=0.5\columnwidth]{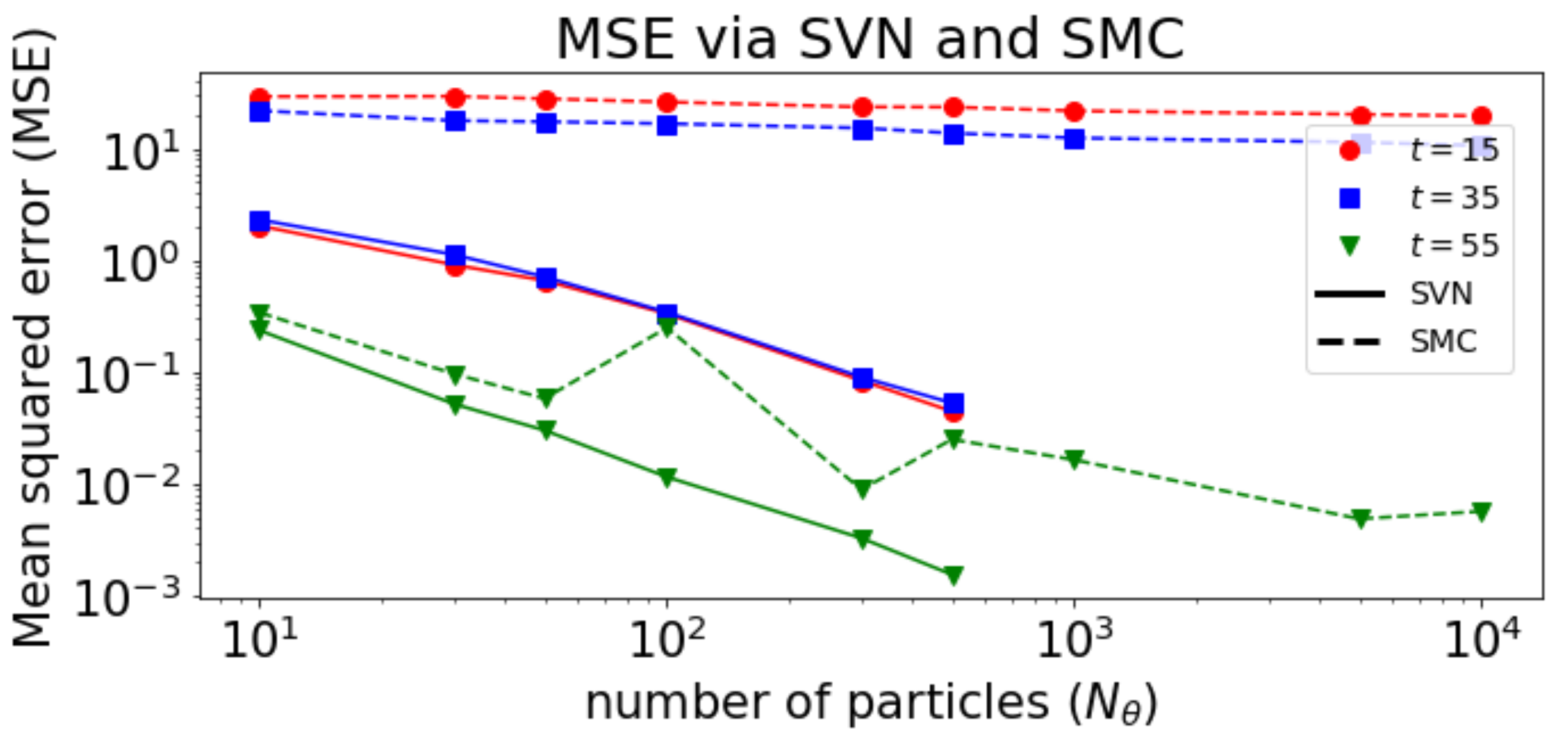}}
		\caption{\small SVN shows a lower mean squared error compared to SMC for lower $N_\theta$.}
		\label{fig:MSE comparison}
	\end{center}\vspace{-1em}
\end{wrapfigure}

\paragraph{Synthetic data. } In order to provide a more direct quantitative comparison, a synthetic Hawkes trajectory of 60 events was constructed with a changepoint every 10th event (see appendix \ref{appendix:synthetic} for details). We measure the trace of the covariance matrix of the posterior via MCMC, and then calculate its mean squared error (MSE) via SVN and SMC (details in appendix \ref{appendix:synthetic}). Figure \ref{fig:MSE comparison} shows how the MSE changes as a function of the number of particles $N_\theta$, at three different values of $\tau$. As $N_\theta$ increases, the error of SVN decreases much faster than that of SMC. Even when using $10^4$ particles for SMC, its error at $\tau = 15, 35$ remains three orders of magnitude larger compared to SVN with 500 particles. A video\footnote{\href{https://gfycat.com/blaringforthrightbullfrog}{https://gfycat.com/blaringforthrightbullfrog}} visualizing the changing posterior contours over time confirms SVN's superiority in tracking changes in complex non-Gaussian distributions: whereas SVN's particles accurately define the posterior, the locations of SMC's particles show instabilities and jump between peaks in the distribution, failing to capturate it as a whole. Table \ref{Tab: quantitative results} summarizes quantitative comparisons between the two methods, including false positive and negative rates of changepoint detection, showing a better performance for SVOCD over BOCPD + SMC.

\vspace{-0.5em}
\section{Application to long short-term memory neural networks}\label{sec:LSTM}
In this section, we adopt \textit{Bayesian long short-term memory} (BLSTM, \cite{hochreiter1997long, fortunato2017bayesian, van2017bayesian}) neural networks as a predictive model. 
%
We note that although frequentist's methods have been proposed to detect changepoints in LSTM \cite{guo2016robust}, this is, to the best of our knowledge, the first fully-Bayesian online changepoint analysis for LSTM.
To demonstrate the effectiveness of SVOCD, we also apply BOCPD using a similar SMC method as in Section 3 to perform the parameter sampling step. 

\paragraph{BLSTM.} 
Here we describe our Bayesian approach to LSTM. For simplicity, we use a time series of scalar data points $y_m\in \R$; our analysis can readily be extended to more general cases.
Consider $\cF_{y_{\tau:m}}(\theta)$ to be the output of the forward pass of a many-to-one LSTM, trained on data $y_{\tau:m}$ and evaluated at $\theta\in\R^d$. The latter contains all the unknown weights and biases in the architecture of the network. If the training set of LSTM is empty, the network will output the bias of the last layer. It will prove to be useful to also construct the corresponding many-to-many LSTM defined by $\bF_{\tau:m} = [\cF_{\emptyset}, \cF_{y_{\tau}}, \cF_{y_{\tau:\tau+1}}\dots, \cF_{y_{\tau:m}}]$.

When $\tau = \tau_{m+1}$, the output of $\cF_{y_{\tau_{m+1}:m}}(\theta)$ is considered a noisy prediction of the data point $y_{m+1}$: 
\begin{equation}\label{eq:LSTMpredModel}
y_{m+1} = \cF_{\tau_{m+1}:m}(\theta) + \sigma \xi\,,
\end{equation}
where $\xi\sim \cN(0,1)$ and $\sigma > 0$. Equation \eqref{eq:LSTMpredModel} is equivalent to defining the predictive likelihood
\begin{equation}\label{eq:LSTMpredLikelihood}
p(y_{m+1}|Y_{\tau_{m+1}},\theta) = \cN(\cF_{\tau_{m+1}:m}(\theta), \sigma^2)(y_{m+1})\,,
\end{equation}
where the right-hand-side of \eqref{eq:LSTMpredLikelihood} denotes a Gaussian density evaluated at $y_{m+1}$ with mean $\cF_{\tau_{m+1}:m}(\theta)$ and variance $\sigma^2$. From equation \eqref{eq:LSTMpredLikelihood} and the relation between $\cF_{\tau_{m+1}:m}$ and $\bF_{\tau_{m+1}:m}$, we find that the likelihood is given by
\begingroup\makeatletter\def\f@size{9.5}\check@mathfonts
\def\maketag@@@#1{\hbox{\m@th\normalsize\normalfont#1}}
\begin{equation}\label{eq:LSTMlkd}
p(y_{\sss\tau_{\sss m+1}:m}|\theta, \tau_{\sss m+1})\! =\! \cN(\bF_{\hspace{-0.05cm}\sss\tau_{\sss m+1}:m}(\theta), \sigma^2 I)(y_{\sss \tau_{\sss m+1}:m})\,.
\end{equation}
\endgroup
Finally, we define a Gaussian prior $p(\theta) = \cN(\mu_0, \sigma_0^2 I)$ and use Bayes' theorem as in \eqref{eq:postDistr} to obtain the posterior distribution.

\paragraph{Backprop and Fisher Information.}
Here we describe how to calculate the Fisher Information of the log-likelihood, which will be used in the SVN algorithm. In deterministic LSTM, backpropagation consists of a gradient descent step which runs backwards, from the last data point to the first. In a Bayesian framework, this corresponds to calculating the gradient of the log-likelihood density:
\begin{equation}
\nabla\log p(y_{\tau_{m+1}:m}|\theta, \tau_{m+1}) = \frac{1}{\sigma^2} \sum\nolimits_{i=\tau_{m+1}-1}^{m-1} \nabla_\theta \cF_{\tau_{m+1}:i}(\theta)^\top (\cF_{\tau_{m+1}:i}(\theta) - y_{i+1})\,.
\end{equation}
Given the Gaussian error assumption in \eqref{eq:LSTMlkd}, the Fisher Information of the likelihood is given by
\begingroup\makeatletter\def\f@size{9}\check@mathfonts
\def\maketag@@@#1{\hbox{\m@th\normalsize\normalfont#1}}
\begin{equation}\label{eq:LSTMHL}
 H_{\sss \cL, \tau_{m+1}}(\theta) := \frac{1}{\sigma^2} \,\, \sum\nolimits_{\sss i=\tau_{m+1}-1}^{\sss m-1} \!\!\!\! \nabla_\theta \cF_{\sss \tau_{m+1}:i}(\theta)^\top\nabla_\theta \cF_{\sss \tau_{m+1}:i}(\theta)\,.
\end{equation}
\endgroup
The Hessian $H_{\pi, \tau_{m+1}}$ of the log-posterior density can now be approximated as in \eqref{eq:Hpi}.

\begin{figure}[h!]
	\vskip 0.2in
	\begin{center}
		\centerline{\includegraphics[trim = 0cm 1cm 0cm 2cm, width=1.0\textwidth, height = 3.5cm]{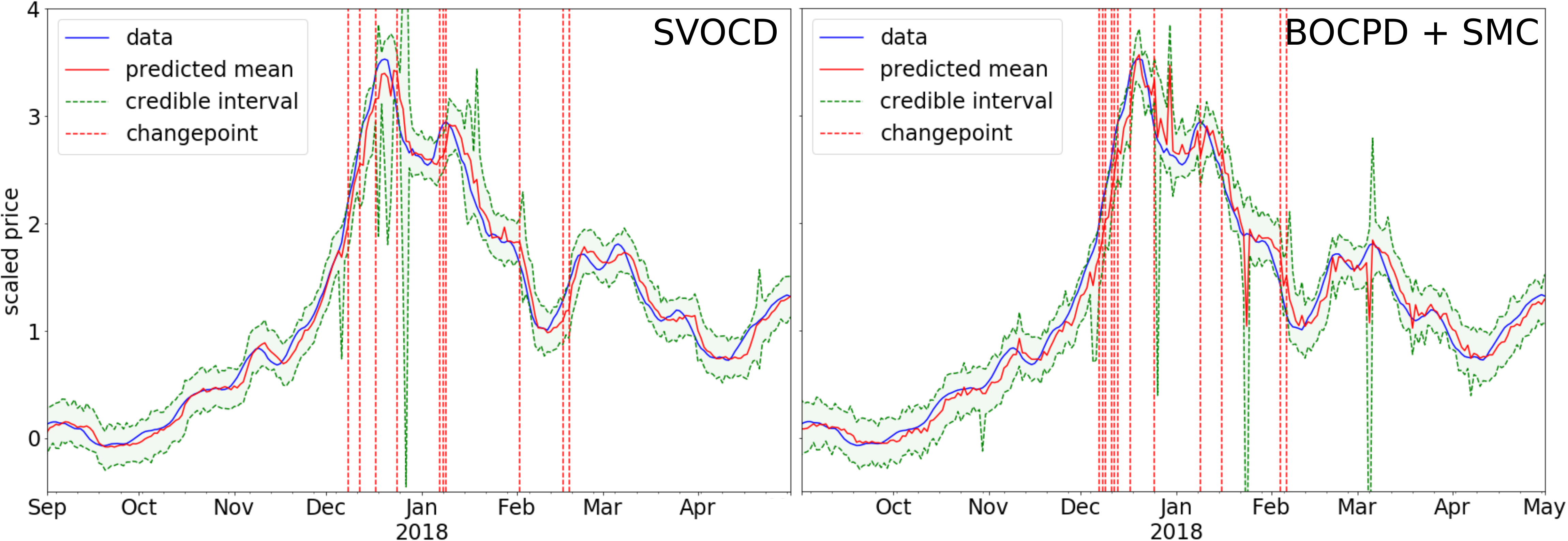}}
		
		\caption{\small Changepoint detection on bitcoin data: (left) SVOCD with BLSTM model; (right) BOCPD + SMC with BLSTM model. BODPC + SMC has more difficulty adapting to changes in trend. SVOCD and BOCPD + SMC were simulated using $N_\theta = 30$ and $N_\theta = 100$ particles, respectively. 100 iterations were used. }
		\label{fig:bitcoin changepoints}
	\end{center}
	\vskip -0.2in
\end{figure}

\paragraph{Application: bitcoin price.}
Bitcoin is a cryptocurrency created in 2009 whose value has fluctuated wildly in the last few years. 
We apply SVOCD and BOCPD + SMC with a BLSTM model to data on bitcoin price evolution (Figure 2 in the supplement), with $\theta\in\R^{64}$. We use a standard Gaussian $\cN(0,I)$ as a prior and a noise level $\sigma = 0.1$ in the likelihood. The hazard rate and the number of predictive samples were fixed at $H_m = 1000$ and $N_y = 100$, respectively.  

Figure \ref{fig:bitcoin changepoints} shows that SVOCD starts to detect changepoints from December 2017 due to a large increase in stock price. After the all-time peak, as the price starts to decrease steeply, another changepoint is detected. Various others are found corresponding to large fluctuations in price. BOCPD + SMC detects changepoints in similar locations, though the increased number of changepoints in the rising phase indicates a difficulty in adapting to changes in trend. In addition, the predicted mean is rougher and less accurate than the one produced by SVOCD, despite using more particles in the simulation.



\section{Conclusion}\label{sec:conclusion}
In this work we introduced SVOCD, a fully-Bayesian method that combines BOCPD and SVN to detect changepoints both online and accurately. 
We successfully applied SVOCD to novel and challenging applications, namely Hawkes processes and LSTM neural networks on WannaCry and Bitcoin real data sets, respectively. A quantitative comparison between SVN and SMC shows that SVN, given its transport nature, is able to carry forward the current estimation of the posterior density which leads to more accurate estimations compared to SMC, even when the number of particles used for SMC is an order of magnitude higher. Further comparisons between SVOCD and BOCPD + SMC, using synthetic data with known changepoints, showed that the former has lower false positive and false negative rates. Because SVOCD samples from the correct posterior, it is able to quickly adapt to changes in trends, to return informative changepoints and to avoid false positives.


\section{Acknowledgements} 
This work was fully sponsored by G-Research. We thank the InfoSecurity division of G-Research for their support. In particular, we would like to acknowledge David Thomas for facilitating the project and Antoine Vianey-Liaud for setting up the WannaCry data set. 

\appendix
\section{Appendix: SVN for Bayesian LSTM}\label{appendix:BLSTM}
In this section, we validate the use of SVN on a Bayesian LSTM model in order to sample correctly from a posterior distribution. In addition, we provide evidence that the prediction given by a Bayesian LSTM is substantially better than the one using a corresponding regularized LSTM.

We use the following simple test case: the data is generated by a noisy sinusoidal signal:
\[ y_j = \sin(j) + \xi\,,\]
where $j = 0,\dots, 50$ are time indices and $\xi\sim \cN(0, 0.15 ^ 2)$ is Gaussian noise. We attempt to reconstruct the data $y_{1:50}$ and its uncertainty by using the Bayesian LSTM model described in the paper. We set a Gaussian prior $p(\theta) = \cN(0,1)$. We further assume a likelihood of the form $\cN(\cF_{1:49}(\theta), 0.3 ^ 2)$,
where $\cF_{1:49}(\theta)$ is the output of the forward pass of a many-to-many LSTM trained on $y_{1:49}$ and evaluated at parameters $\theta \in \R^{64}$. In order to train the BLSTM model, we use SVN with 30 particles initialized around the MAP of the distribution and run it for 100 iterations. 

\begin{figure}[h!]
	\begin{center}
		\centerline{ \includegraphics[width=0.7\columnwidth]{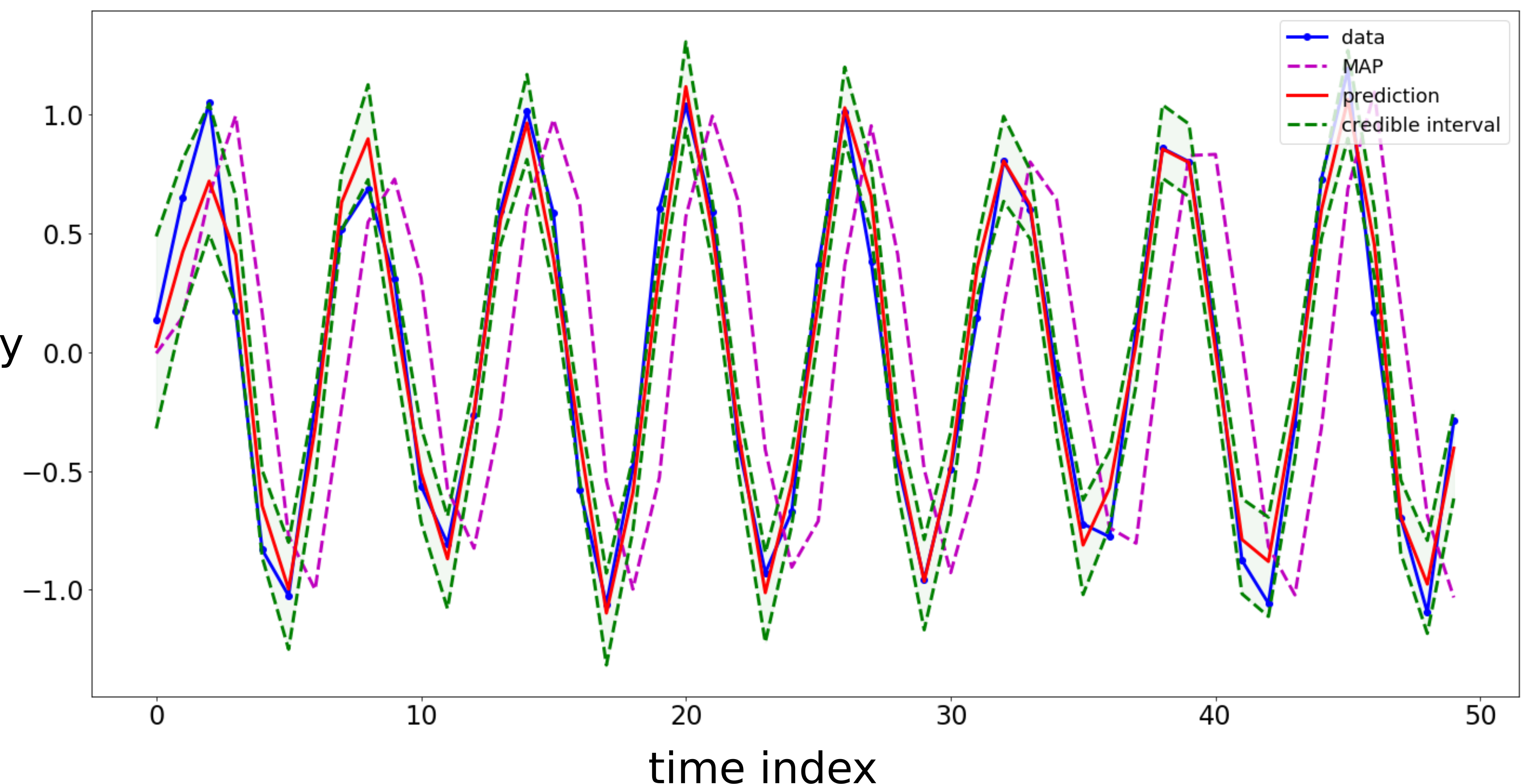}}
		\caption{BLSTM trained with SVN on sinusoidal signal}
		\label{fig:BLSTMsin}
	\end{center}
	\vskip -0.2in
\end{figure}

Figure \ref{fig:BLSTMsin} displays the results of our simulation. The blue line is the real data. The red line is the average of the particles representing our prediction. The green shaded area represents a $95\%$ credible interval around the mean. We can see how the red prediction captures the sinusoidal motion of the signal, while the uncertainty of the signal is well represented by the green area. In contrast, the dashed magenta line is the mode of the distribution, i.e.~the estimator that would be returned by a deterministic regularized LSTM. We can see very clearly that the mode of the distribution overweights the importance of the last observations: at every stage the magenta line almost exactly replicates the previous observation.

In conclusion, we find that: i) SVN is able to correctly represent the posterior distribution, and ii) a Bayesian framework is superior to a deterministic one: it allows us to calculate the average of the posterior distribution, leading to much better predictions compared to using the mode found by deterministic models.

\section{Appendix: Bitcoin price changepoint detection}\label{appendix:bitcoin}

\begin{figure}[h!]
	\begin{center}
		\centerline{ \includegraphics[width=0.6\textwidth]{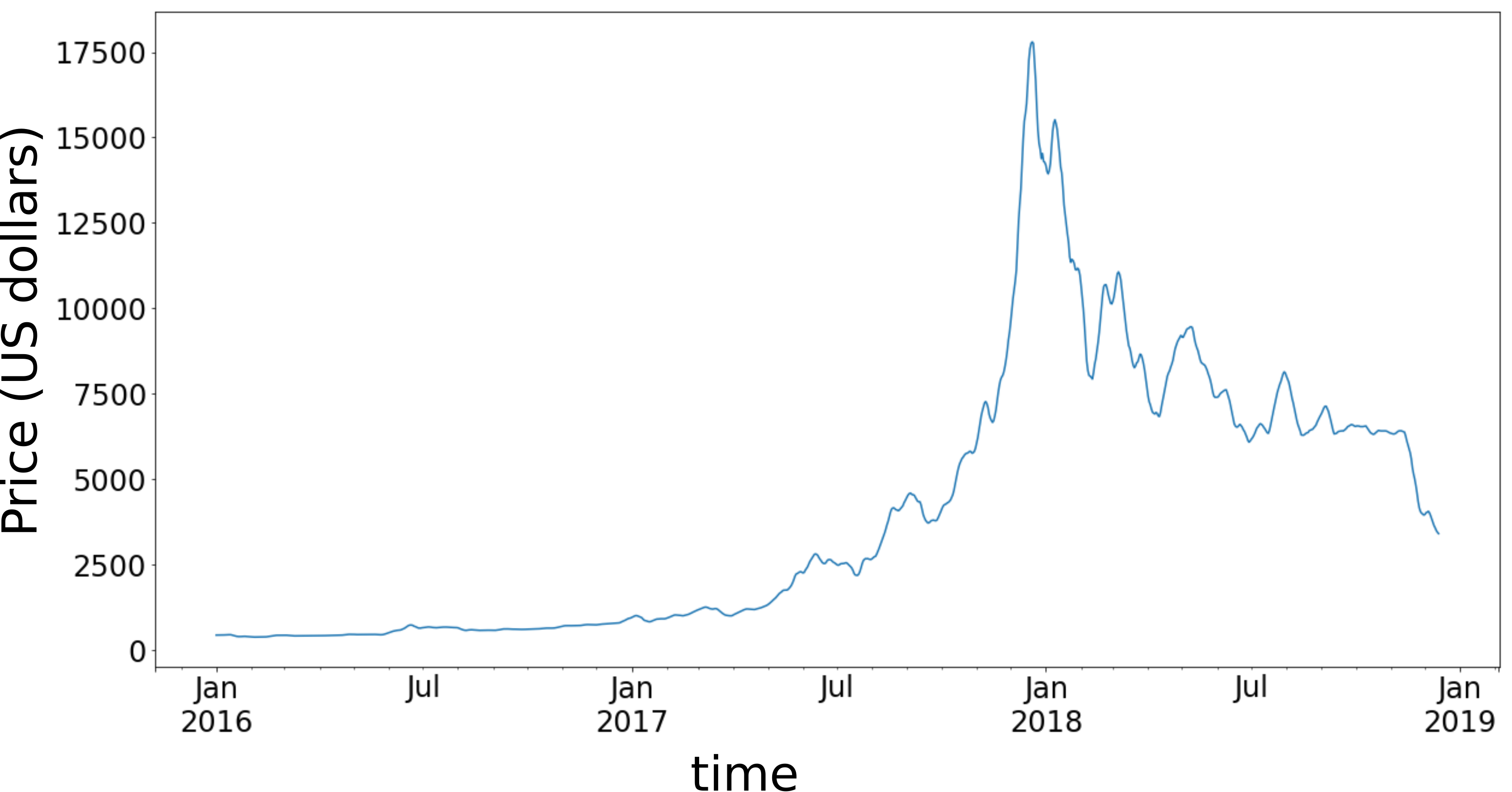}}
		
		\caption{\small Bitcoin price evolution in US dollars.}
		\label{fig:bitcoin_data}
	\end{center}
\end{figure}

Figure \ref{fig:bitcoin_data} shows the weekly rolling-averaged data of the evolution of bitcoin price from the beginning of 2016 to the 13th of December 2018. Price started at $998\$$ in 2017 and rose to $13,412.44\$$ on the 1st of January 2018, with an all-time peak of $19,666\$$ on the 17th of December 2017. From that point on, the price fluctuated downwards up to $3,690\$$ at the end of 2018, about $81\%$ down from the all-time peak.

Figure \ref{fig:bitcoin changepoints} shows the results of SVOCD in the region where the price dynamics reaches its all-time peak and then suddenly drops (outside this region, predictions are very stable and no changepoints are detected). 

\section{Appendix: Synthetic data comparison}\label{appendix:synthetic}
We constructed 60 data points $y_{1:60}$ where each of these observations is generated by a Hawkes process driven by a parameter $\theta = [\mu, \gamma, \delta] \in \R^3$. A changepoint is inserted after every 10th observation, i.e. changepoints occur at $\tau = 10, 20, \dots, 60$. Between these changepoints, the generating parameter $\theta$ alternates between two states, namely $\theta_1 = [-1,-2,1]$ and $\theta_2 = [2,4,0]$. 

\begin{figure}[h!]
	\begin{center}
		\centerline{ \includegraphics[width=1.0\columnwidth]{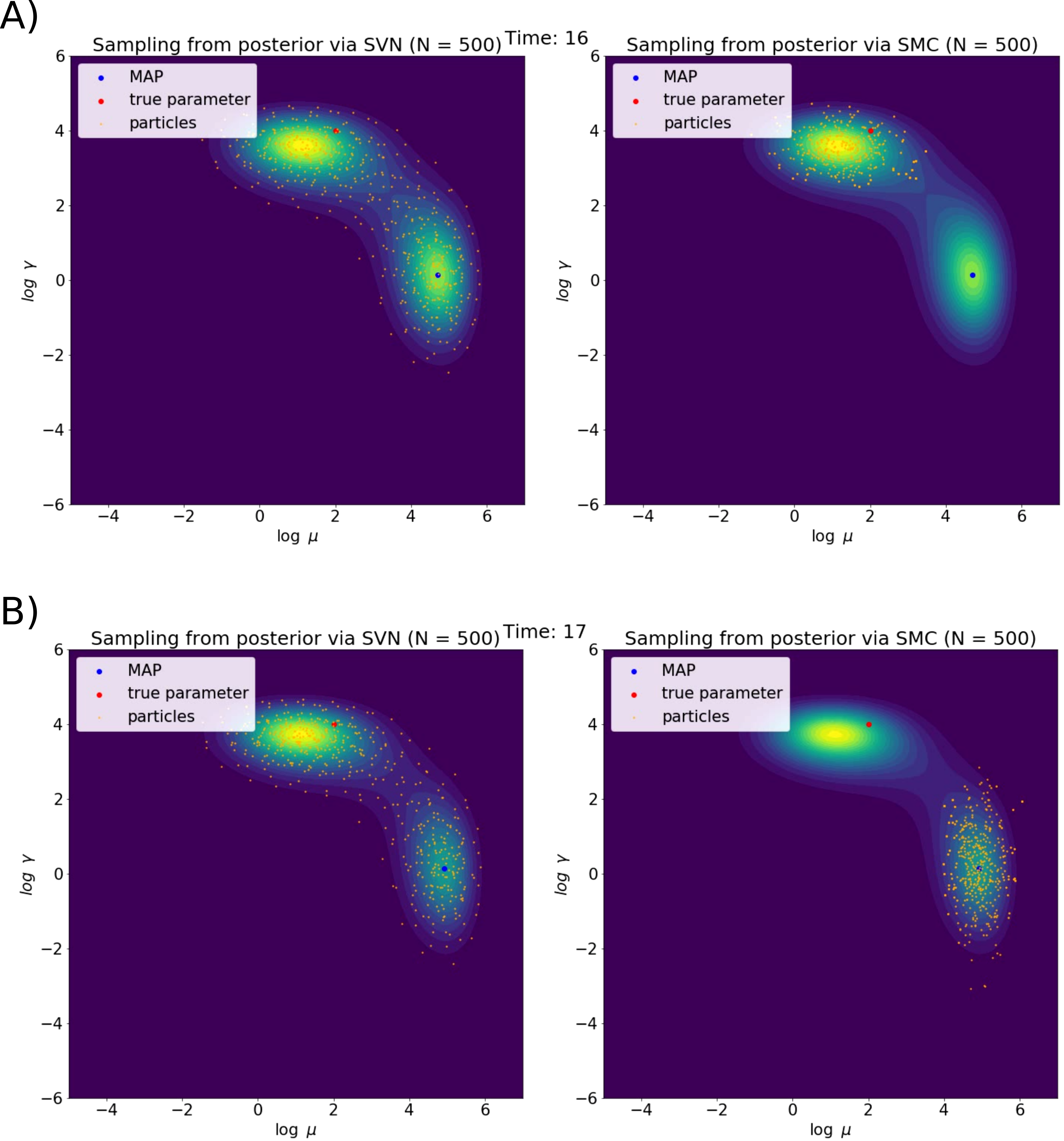}}
		\caption{\textbf{SVN more accurately estimates the posterior density.} We visualize the changing posterior contours, as well as the particle positions of SVN and SMC, at time points A) $\tau = 16$ and B) $\tau = 17$. While SVN's particles accurately describe the posterior, SMC's particles tend to jump from one peak to another without properly describing the entire distribution. These plots were generated using the synthetic data as described above.}
		\label{fig: snapshots video}
	\end{center}
\end{figure}

We set a standard Gaussian prior and a Hawkes process likelihood \eqref{eq:hawkes_llkd}. The goal is to sequentially retrieve samples from the posteriors $p(\theta|Y_{\tau_{m+1}})$, for $m=0,1,\dots,59$. We do this via both SVN and SMC ($N = 500$ particles) and compare their performance. Figure \ref{fig: snapshots video} shows snapshots of the video\footnote{\href{https://gfycat.com/blaringforthrightbullfrog}{https://gfycat.com/blaringforthrightbullfrog}} we made to visualize the performance of SVN and SMC on sequentially tracking the changes of the posterior over time. From these snapshots, we clearly see that SMC fails to accurately describe the posterior density when the geometry of the density is highly non-Gaussian.

In order to produce Figure \ref{fig:MSE comparison}, the same synthetic data was used. First, at each time step, we measure the trace of the covariance matrix of the posterior distribution calculated via a random walk MCMC with $5\times 10^5$ steps. We then also retrieve posterior samples via SVN and SMC for a range of values of $N$ ($N=10, 30, 50, 100, 300, 500, 1000$ for SVN and $N=10, 30, 50, 100, 300, 500, 1000, 5000, 10000$ for SMC). For each value of $N$, 30 runs are performed to measure the mean and variance of the mean squared error (MSE), calculated as the mean squared difference between the traces of the covariance matrix via the random walk and via SVN or SMC. In Figure \ref{fig:MSE comparison}, we can clearly see a better convergence for SVN compared to SMC, confirming the better performance of SVN as observed in the video. 

Furthermore, on the same synthetic data, we perform 30 runs of both SVOCD and BOCPD + SMC to measure mean and standard deviation of false positive and false negative changepoint rates for both methods. Table \ref{Tab: quantitative results} shows that, as expected from the better convergence results discussed previously, SVOCD achieves smaller means for both false positive and false negative rates.

\clearpage



\printbibliography
\nocite{*}

\end{document}